\documentclass[fontsize=12pt]{ceurart}

\usepackage{listings}

\usepackage{booktabs}
\usepackage{array}    % For column types
\begin{document}

\copyrightyear{2024}
\copyrightclause{Copyright for this paper by its authors.
  Use permitted under Creative Commons License Attribution 4.0
  International (CC BY 4.0).}

\conference{CHR 2024: Computational Humanities Research Conference, December 4–6, 2024, Aarhus, Denmark}

\title{Locating the Leading Edge of Cultural Change}

\author[1]{Sarah Griebel}[%
orcid=0009-0009-6909-0319,
email=sarahg8@illinois.edu
]
\address[1]{School of Information Sciences, University of Illinois, Urbana-Champaign, USA}
\address[2]{Nanyang Technological University, Singapore}

\author[1]{Becca Cohen}[%
email=rscohen2@illinois.edu
]

\author[1]{Lucian Li}[%
orcid=0000-0002-9462-6527,
email = zilul2@illinois.edu
]

\author[2]{Jaihyun Park}[%
orcid=0000-0001-6217-7192,
email=jay.park2@ntu.edu.sg
]

\author[1]{Jiayu Liu}[%
email=jiayu13@illinois.edu
]

\author[1]{Jana Perkins}[%
email=jmp12@illinois.edu]

\author[1]{Ted Underwood}[%
orcid = 0000-0001-8960-1846,
email = tunder@illinois.edu]
\cormark[1]

\date{\today}

\cortext[1]{Corresponding author.}

\begin{abstract}
  Measures of textual similarity and divergence are increasingly used to study cultural change. But which measures align, in practice, with social evidence about change? We apply three different representations of text (topic models, document embeddings, and word-level perplexity)  to three different corpora (literary studies, economics, and fiction). In every case, works by highly-cited authors and younger authors are textually ahead of the curve. We don't find clear evidence that one representation of text is to be preferred over the others. But alignment with social evidence is strongest when texts are represented through the top quartile of passages, suggesting that a text's impact may depend more on its most forward-looking moments than on sustaining a high level of innovation throughout.
\end{abstract}

\date{}
\maketitle

\section{Introduction}

A growing body of scholarship seeks to understand cultural change by measuring the way individual texts precede or lag corpus-level trends. 

Different disciplines have framed this problem differently. Fields like bibliometrics measure novelty by comparing an article to past precedent, and ask how well novelty predicts impact as measured by citations \cite{ZHANG2021101140}. By contrast, some computational social scientists are less interested in divergence from the past than in anticipation of the future. In Vicinanza et al. 2022, for instance, a text’s “prescience,” or anticipation of future change, is used to identify social locations where new patterns tend to emerge~\cite{Vicinanza2022}. It is also possible to combine both approaches, and study a text’s relationship to past and future at once. Barron et al. 2018 measures a text’s divergence from the past (“novelty”) and subtracts divergence from the future (“transience”)---producing a measure of durable innovation they call “resonance'' ~\cite{Barron2018}.

Models of textual change have also relied on radically different representations of text, ranging from lexical topic models in~\cite{Barron2018} to a deep-learning model of sentences in~\cite{Vicinanza2022}. Plausible \emph{a priori} arguments can be made for all of these methods. In this paper we will try to provide empirical evidence about best practices.

To empirically assess methods of measuring textual change, of course, we need some kind of ground truth about a text's divergence from the past (or similarity to the future). This is not a topic where absolute ground truth is available. In fact, researchers measure innovation textually because they have reason to suspect that social evidence will be unreliable here. So instead of relying on a single unimpeachable source of social evidence, we may have to combine several.

For instance, bibliometricians have repeatedly confirmed that innovation does correlate with publicity~\cite{ZHANG2021101140, bornmann2019measurenovelty}. Works that introduce new language, or cite new combinations of sources, tend to attract more attention and receive more citations themselves. So we could use citation frequency as one signal that a text was on the leading edge of change. 

But we also have reason to suspect that using publicity as a measure of innovation will overrate already-prominent writers, who tend to receive more attention through the “Matthew effect”~\cite{Merton_1968}. Young writers are cited less frequently. And yet many ideas and locutions must emerge first in young writers, since cohort succession is one driver of cultural change~\cite{Meisel2013, Miller1996, Underwood_2022}.

The tension between these two forms of social evidence gives us leverage on the problem. If we can find a measure of a text’s relation to change that aligns well with youth but also with citation and prominence, we will have validated our measure against two independent variables, suggesting that it describes “the leading edge of cultural change” in a relatively broad and robust sense.

The documents we consider include journal articles drawn from literary studies and economics, as well as a collection of English-language fiction ranging from 1890 to 2000. In both cases, we have all or most of the documents in full text, so we can compare Transformer-based models to older strategies of lexical modeling.

Our experiment supports several inferences about best practices for measuring change. For instance, is a text’s relation to the past or the future more informative? When do Transformer-based models outperform lexical ones? Should texts always be considered as wholes, or might it be more meaningful to represent them through their most innovative parts?

\section{Data}

We modeled the impact of textual innovation using three datasets. Two datasets contained academic journal articles from the fields of literary studies and economics---fields selected because we expect their rhetorical and citation practices to diverge significantly. One contained English-language fiction.

\subsection{Academic journals}

Journals were selected for longevity and influence in the field. Journals with longer lifespans were prioritized, as this would ensure relative stability across the corpus.

The literary studies dataset contains a corpus of 40,407 full text academic articles from seven journals. The economics dataset contains 43,081 articles from eight journals. Texts were obtained through JSTOR~\cite{JSTOR_DfR}. Consult Appendix E for a full list of journal titles. Citation counts were gathered from Semantic Scholar~\cite{Kinney2023TheSS}. Authors' years of birth were inferred through a mixture of manual checking and matching to VIAF, which gave us age at publication for 2,646 articles in literary studies (see Appendix D for our methods of inference). Youth and citation frequency are negatively correlated in this corpus.

\subsection{Fiction}

We gathered 8,918 works of English-language fiction distributed approximately evenly across time from 1890 through 2000. The first and last 10\% of each book was discarded to avoid mixing fiction with introductions, advertisements, and other nonfiction paratext. Of our 8,918 books, only 7,304 are in full text; since we only produced embeddings of these books, the embedding method had a slight disadvantage on the fiction corpus~\cite{jett2020hathitrust}.

We drew information about authors’ years of birth from Underwood et al. 2022, which gave us author ages for 3,272 works in the period we were analyzing~\cite{Underwood_2022}.

We also created a subset of “critically discussed” works by finding the titles and authors of our fiction corpus in our literary studies corpus. This group of 463 books was compared to a contrast set with the same distribution across time, but never mentioned in that corpus.

\section{Methods}

We measured both divergence from the past (which following Barron et al. 2018 we call “novelty”) and divergence from future documents (“transience”). But most of the results below are based on the composite quantity they call “resonance” (novelty minus transience). To avoid any suggestion of causality we call this quantity ``precocity.'' A text with high precocity simply “looks later than” peers published in the same year. We calculate these quantities using three different representations of texts.

\subsection{Topic models}

We topic modeled our corpora using the implementation of LDA in MALLET, and divided documents into chunks of at least 512 tokens~\cite{McCallum2002, blei2003latent}. For more detail, see Appendix A. We compared documents by measuring Kullback-Leibler divergence on topic distributions, following Barron et al. 2018 \cite{Barron2018}. 

\subsection{Document embeddings}

Topic models are limited to lexical evidence. It seemed plausible that neural document embeddings, compared via cosine distance, might capture a richer representation of text. We experimented with several different embedding strategies. Off-the-shelf embeddings performed very poorly, even if they were at the top of the leaderboard for contemporary applications. Fine-tuning using the sentence Transformers library was necessary to produce embeddings more suited to the specialized subject matter and temporal range (1890--2017) of this experiment~\cite{reimers-2019-sentence-bert}. See Appendix B for details of our fine-tuning strategy.

\subsection{Perplexity}

Vicinanza et al. measure a quantity they call “prescience,” which is calculated by comparing the mean perplexity of a document’s sentences in two models---one trained via masked language modeling on its own period, and one trained on a future period. Sentences that have lower perplexity in the future (which become more probable in the future) will have high prescience. In bibliometrics, a loosely similar method has been used to compute novelty~\cite{Vicinanza2022, 10.1371/journal.pone.0284567, 10.1371/journal.pone.0254034}.

We tested Vicinanza's definition of prescience on our corpora, but found that we got much greater predictive power by using an expanded version of the method that included both past and future. Instead of subtracting future perplexity from a document’s perplexity at time of publication, we subtract it from perplexity calculated in the past.

\begin{equation}
\text{precocity} = 2 \cdot \frac{\text{perplexity}_{\text{past}} - \text{perplexity}_{\text{future}}}{\text{perplexity}_{\text{past}} + \text{perplexity}_{\text{future}}}
\end{equation}

This measures not just anticipation of a specific future period, but a quality of being “ahead of the curve,” where the curve is inferred from the whole time window around publication of a text. For further details see Appendix F.

\subsection{Details of precocity calculation}

Documents were divided into chunks for all three of these strategies, and chunks were characterized individually. For the first two methods this meant that each chunk was compared to all the other document chunks in the preceding (and following) 20 years. Perplexity relied on models that characterize a 12-year period, so direct chunk-to-chunk comparisons were not required. The full span from the ``past” model to the ``future” model was 36 years, rendering the scale of the perplexity calculation comparable to the 41-year span of the other two methods.

It is certainly possible to characterize a document through the mean precocity of its chunks. But an argument can also be made that what matters, socially, is often not the average tenor of a document, but its most surprising and forward-looking moment. For this reason we also tested an alternate strategy that characterized documents by selecting the top 25\% of their chunks with highest precocity, and taking the mean of those values.

An alert reader will anticipate that questions of circularity might emerge when texts quote each other or were written by the same author. See Appendix C for our solution to these problems. In practice these effects were very small; excluding or leaving in texts that quote each other made almost no difference.

\subsection{Regression strategy}

We assess the explanatory power of precocity through a multiple linear regression that includes terms for precocity, precocity squared, and novelty (which gives the regression leverage to separate the components of precocity that refer to the past or to the future). Date of publication is also present as a control variable.

\section{Results}
We’ll begin with a quick overview of the variance explained when six different methods of text analysis are applied to predicting five social variables.

\begin{table}[ht]
\centering
\caption{R\textsuperscript{2} for different representations of text, on different social variables. 0.25 indicates that documents were represented by the quartile of passages with highest precocity; 1.0, represented by all passages.}
\label{tab:your_table}
\begin{tabular}{@{}l *{6}{p{1.55cm}} @{}}
\toprule
        & Topics, 0.25 & Embeds, 0.25 & Perplexity, 0.25 & Topics,\hspace{0.3cm}  1.0 & Embeds, 1.0 &  Perplexity, 1.0 \\ \midrule
Citations, literary scholars & \textbf{.082}            & .070             & .057            & .067            & .049           & .041           \\ \midrule
Age, literary scholars      & .033            & \textbf{.035}            & .024            & .026             & .034           & .024           \\ \midrule
Critical discussion, fiction         & \textbf{.071}            & .011            & .013                & .033            & .009           & .012               \\ \midrule
Age, fiction writers        & .051            & \textbf{.083}            & .014                & .049            & .076            & .014          \\ \midrule
Citations, economists  & .063  & .029   & \textbf{.096}        & .040   & .018 & .063  \\ \bottomrule
\end{tabular}
\end{table}

As we predicted, textual innovation is associated both with prominence and with youth (even though a text's prominence is anti-correlated with youth in this data). The best-performing methods were able to explain 7-9\% of the variance in prominence (e.g., citation count) simply by identifying works that were (textually) ahead of the curve---more like the future than the past. 

It is difficult to say whether explaining 7-9\% of social variance is good performance, because we don't know how much of a work's prominence is really determined by innovation--and how much by factors like institutional prestige. Some research suggests that the answer varies from one discipline to another~\cite{ZHANG2021101140}. It nevertheless seems reasonable to take social variance explained as a heuristic to choose between methods---for while we don’t know the real effect size, it is unclear how significant effects larger than the real one would be produced.

So what did we learn about measuring precocity? The clearest lesson is that the signal tended to be 
strongest when we measured documents “at their most forward-looking,'' by averaging the 25\% of passages with the highest precocity scores. In all of the tests we ran, this method aligned better with social evidence than a method that averaged all passages. One might infer that citations---and more surprisingly, critical references to fiction---are often motivated by innovations expressed in a relatively small part of a text.

Second, on examining regression coefficients, transience (difference from the future) provides slightly stronger evidence of failure than novelty (difference from the past) provides evidence of success. The difference between these variables was not huge, however, and there was nothing to be gained by discarding information about the past. The original perplexity method in Vicinanza et al. 2022, which only included information about the future, achieved r\textsuperscript{2} less than half as large as the improved method we describe in the table above. Comparing texts only to the past, or only to the future, would admittedly make it easier to use causal language. Precocity, which characterizes a text in relation to a whole time window around its publication, is hard to interpret causally. But if causal explanation is not being claimed, there is no reason not to use both time arrows at once.

All three representations of text (topic models, embeddings, and perplexity) performed well in some cases. Topic models seemed to predict prominence well, while embeddings performed well on age---but we don't have enough data points to generalize. If any conclusion can be drawn here, it might be ``a dog that doesn't bark.'' We found no evidence that neural models of text systematically outperformed lexical models. On the contrary, lexical topic models displayed consistently strong performance across tasks and corpora.

\section{Discussion}
We found clear alignment between textual measures of precocity and two independent kinds of social evidence that we expected to align with change (prominence and authorial youth). There is no absolute ground truth in this domain, but statistically significant relationships across three corpora do increase our confidence that text analysis can locate a leading edge of cultural change.

We also consistently found a better fit with social evidence when we represented documents through the 25\% of passages with highest precocity. It seems likely that significant innovations are often concentrated in a small portion of an article or work of fiction.

However, we also found that precocity can be measured in different ways, which do not always agree with each other. Since change is taking place in a space that has multiple dimensions, the same text can be at the leading edge on one axis and lagging behind on another. Measures of similarity based on a topic model seemed to excel at predicting citations and public reputation. Transformer-based embeddings may be better at catching tacit signals of generational style.

Since topic modeling is an older representation of text, its strong performance overall may require discussion. We don't have a confident answer here, but for what it’s worth, topic models are explicitly designed to factor a corpus into latent variables. Document embeddings, by contrast, don’t have any representational goal at the corpus level. The embeddings we used are tuned contrastively, using the Sentence Transformers library~\cite{reimers-2019-sentence-bert}. But that process is not guaranteed to model the corpus in a principled way---which might be a disadvantage in an experiment that seeks to measure a document’s relation to corpus-level trends.

At least for now, researchers wrestling with questions about textual change are well advised to evaluate the performance of a principled lexical model as a baseline and confirm that embeddings do actually improve on it before relying on embeddings alone. It is not safe to assume that a model will perform better simply because it captures information about word order. 

\section{Public data and code}

Data and code for this project are available on GitHub:
\href{https://github.com/IllinoisLiteraryLab/novelty/tree/main}{\nolinkurl{https://github.com/IllinoisLiteraryLab/novelty/tree/main}}, and will also be archived on Zenodo.

\section{Contribution statement}
Authors are listed alphabetically here.

Conceived and designed the analysis: Becca Cohen, Sarah Griebel, Lucian Li, Jiayu Liu, Jaihyun Park, Jana Perkins, Ted Underwood

Wrote the paper: Becca Cohen, Sarah Griebel, Lucian Li, Jiayu Liu, Jaihyun Park, Jana Perkins, Ted Underwood

Collected the data: Becca Cohen, Sarah Griebel, Lucian Li, Ted Underwood

Contributed data or analysis tools: Becca Cohen, Sarah Griebel, Lucian Li, Ted Underwood

Performed the analysis: Sarah Griebel, Ted Underwood

\begin{acknowledgments}
  This work made use of the Illinois Campus Cluster, a computing resource that is operated by the Illinois Campus Cluster Program (ICCP) in conjunction with the National Center for Supercomputing Applications (NCSA) and which is supported by funds from the University of Illinois at Urbana-Champaign---specifically, through the Illinois Computes program. This work also used the Delta system at the National Center for Supercomputing Applications through allocation xras-ncsa-72 from the Advanced Cyberinfrastructure Coordination Ecosystem: Services \& Support (ACCESS) program, which is supported by National Science Foundation grants \#2138259, \#2138286, \#2138307, \#2137603, and \#2138296. Some fiction data for this project was provided by HathiTrust Digital Library~\cite{jett2020hathitrust}.
\end{acknowledgments}

%%
%% Define the bibliography file to be used
\bibliography{bibliography}

\begin{thebibliography}{22}
\expandafter\ifx\csname natexlab\endcsname\relax\def\natexlab#1{#1}\fi
\providecommand{\url}[1]{\texttt{#1}}
\providecommand{\href}[2]{#2}
\providecommand{\path}[1]{#1}
\providecommand{\DOIprefix}{doi:}
\providecommand{\ArXivprefix}{arXiv:}
\providecommand{\URLprefix}{URL: }
\providecommand{\Pubmedprefix}{pmid:}
\providecommand{\doi}[1]{\href{http://dx.doi.org/#1}{\path{#1}}}
\providecommand{\Pubmed}[1]{\href{pmid:#1}{\path{#1}}}
\providecommand{\bibinfo}[2]{#2}
\ifx\xfnm\relax \def\xfnm[#1]{\unskip,\space#1}\fi
%Type = Article
\bibitem[{Zhang et~al.(2021)Zhang, Xie, and Song}]{ZHANG2021101140}
\bibinfo{author}{X.~Zhang}, \bibinfo{author}{Q.~Xie}, \bibinfo{author}{M.~Song},
\newblock \bibinfo{title}{Measuring the {I}mpact of {N}ovelty, {B}ibliometric, and {A}cademic-{N}etwork {F}actors on {C}itation {C}ount {U}sing a {N}eural {N}etwork},
\newblock \bibinfo{journal}{Journal of Informetrics} \bibinfo{volume}{15} (\bibinfo{year}{2021}) \bibinfo{pages}{101140}. \DOIprefix\doi{https://doi.org/10.1016/j.joi.2021.101140}.
%Type = Article
\bibitem[{Vicinanza et~al.(2022)Vicinanza, Goldberg, and Srivastava}]{Vicinanza2022}
\bibinfo{author}{P.~Vicinanza}, \bibinfo{author}{A.~Goldberg}, \bibinfo{author}{S.~B. Srivastava},
\newblock \bibinfo{title}{{A {D}eep-{L}earning {M}odel of {P}rescient {I}deas {D}emonstrates {T}hat {T}hey {E}merge from the {P}eriphery}},
\newblock \bibinfo{journal}{PNAS Nexus} \bibinfo{volume}{2} (\bibinfo{year}{2022}) \bibinfo{pages}{pgac275}. \DOIprefix\doi{10.1093/pnasnexus/pgac275}.
%Type = Article
\bibitem[{Barron et~al.(2018)Barron, Huang, Spang, and DeDeo}]{Barron2018}
\bibinfo{author}{A.~T.~J. Barron}, \bibinfo{author}{J.~Huang}, \bibinfo{author}{R.~L. Spang}, \bibinfo{author}{S.~DeDeo},
\newblock \bibinfo{title}{Individuals, {I}nstitutions, and {I}nnovation in the {D}ebates of the {F}rench {R}evolution},
\newblock \bibinfo{journal}{Proceedings of the National Academy of Sciences} \bibinfo{volume}{115} (\bibinfo{year}{2018}) \bibinfo{pages}{4607--4612}. \DOIprefix\doi{10.1073/pnas.1717729115}.
%Type = Misc
\bibitem[{Bornmann et~al.(2019)Bornmann, Tekles, Zhang, and Ye}]{bornmann2019measurenovelty}
\bibinfo{author}{L.~Bornmann}, \bibinfo{author}{A.~Tekles}, \bibinfo{author}{H.~H. Zhang}, \bibinfo{author}{F.~Y. Ye}, \bibinfo{title}{Do {W}e {M}easure {N}ovelty {W}hen {W}e {A}nalyze {U}nusual {Combinations of Cited References? A Validation Study of Bibliometric Novelty Indicators Based on F1000Prime Data}}, \bibinfo{year}{2019}. \URLprefix \url{https://arxiv.org/abs/1910.03233}. \href{http://arxiv.org/abs/1910.03233}{{\tt arXiv:1910.03233}}.
%Type = Article
\bibitem[{Merton(1968)}]{Merton_1968}
\bibinfo{author}{R.~K. Merton},
\newblock \bibinfo{title}{The {M}atthew {E}ffect in {S}cience},
\newblock \bibinfo{journal}{Science} \bibinfo{volume}{159} (\bibinfo{year}{1968}) \bibinfo{pages}{56--63}. \DOIprefix\doi{10.1126/science.159.3810.56}.
%Type = Book
\bibitem[{Meisel et~al.(2013)Meisel, Elsig, and Rinke}]{Meisel2013}
\bibinfo{author}{J.~M. Meisel}, \bibinfo{author}{M.~Elsig}, \bibinfo{author}{E.~Rinke}, \bibinfo{title}{{Language Acquisition and Change: A Morphosyntactic Perspective}}, \bibinfo{publisher}{Edinburgh University Press}, \bibinfo{address}{Edinburgh}, \bibinfo{year}{2013}.
%Type = Book
\bibitem[{Miller and Shanks(1996)}]{Miller1996}
\bibinfo{author}{W.~E. Miller}, \bibinfo{author}{J.~M. Shanks}, \bibinfo{title}{{The New American Voter}}, \bibinfo{publisher}{Harvard University Press}, \bibinfo{address}{Cambridge, MA}, \bibinfo{year}{1996}.
%Type = Article
\bibitem[{Underwood et~al.(2022)Underwood, Kiley, Shang, and Vaisey}]{Underwood_2022}
\bibinfo{author}{T.~Underwood}, \bibinfo{author}{K.~Kiley}, \bibinfo{author}{W.~Shang}, \bibinfo{author}{S.~Vaisey},
\newblock \bibinfo{title}{{Cohort Succession Explains Most Change in Literary Culture}},
\newblock \bibinfo{journal}{Sociological Science} \bibinfo{volume}{9} (\bibinfo{year}{2022}) \bibinfo{pages}{184--205}. \DOIprefix\doi{10.15195/v9.a8}.
%Type = Inproceedings
\bibitem[{Burns et~al.(2009)Burns, Brenner, Kiser, Krot, Llewellyn, and Snyder}]{JSTOR_DfR}
\bibinfo{author}{J.~Burns}, \bibinfo{author}{A.~Brenner}, \bibinfo{author}{K.~Kiser}, \bibinfo{author}{M.~Krot}, \bibinfo{author}{C.~Llewellyn}, \bibinfo{author}{R.~Snyder},
\newblock \bibinfo{title}{{JSTOR} - {Data for Research}},
\newblock in: \bibinfo{editor}{M.~Agosti}, \bibinfo{editor}{J.~Borbinha}, \bibinfo{editor}{S.~Kapidakis}, \bibinfo{editor}{C.~Papatheodorou}, \bibinfo{editor}{G.~Tsakonas} (Eds.), \bibinfo{booktitle}{Research and Advanced Technology for Digital Libraries}, \bibinfo{publisher}{Springer Berlin Heidelberg}, \bibinfo{address}{Berlin, Heidelberg}, \bibinfo{year}{2009}, pp. \bibinfo{pages}{416--419}.
%Type = Article
\bibitem[{Kinney et~al.(2023)Kinney, Anastasiades, Authur, Beltagy, Bragg, Buraczynski, Cachola, Candra, Chandrasekhar, Cohan, Crawford, Downey, Dunkelberger, Etzioni, Evans, Feldman, Gorney, Graham, Hu, Huff, King, Kohlmeier, Kuehl, Langan, Lin, Liu, Lo, Lochner, MacMillan, Murray, Newell, Rao, Rohatgi, Sayre, Shen, Singh, Soldaini, Subramanian, Tanaka, Wade, Wagner, Wang, Wilhelm, Wu, Yang, Zamarron, van Zuylen, and Weld}]{Kinney2023TheSS}
\bibinfo{author}{R.~M. Kinney}, \bibinfo{author}{C.~Anastasiades}, \bibinfo{author}{R.~Authur}, \bibinfo{author}{I.~Beltagy}, \bibinfo{author}{J.~Bragg}, \bibinfo{author}{A.~Buraczynski}, \bibinfo{author}{I.~Cachola}, \bibinfo{author}{S.~Candra}, \bibinfo{author}{Y.~Chandrasekhar}, \bibinfo{author}{A.~Cohan}, \bibinfo{author}{M.~Crawford}, \bibinfo{author}{D.~Downey}, \bibinfo{author}{J.~Dunkelberger}, \bibinfo{author}{O.~Etzioni}, \bibinfo{author}{R.~Evans}, \bibinfo{author}{S.~Feldman}, \bibinfo{author}{J.~Gorney}, \bibinfo{author}{D.~W. Graham}, \bibinfo{author}{F.~Hu}, \bibinfo{author}{R.~Huff}, \bibinfo{author}{D.~King}, \bibinfo{author}{S.~Kohlmeier}, \bibinfo{author}{B.~Kuehl}, \bibinfo{author}{M.~Langan}, \bibinfo{author}{D.~Lin}, \bibinfo{author}{H.~Liu}, \bibinfo{author}{K.~Lo}, \bibinfo{author}{J.~Lochner}, \bibinfo{author}{K.~MacMillan}, \bibinfo{author}{T.~C. Murray}, \bibinfo{author}{C.~Newell}, \bibinfo{author}{S.~R. Rao}, \bibinfo{author}{S.~Rohatgi}, \bibinfo{author}{P.~Sayre},
  \bibinfo{author}{Z.~Shen}, \bibinfo{author}{A.~Singh}, \bibinfo{author}{L.~Soldaini}, \bibinfo{author}{S.~Subramanian}, \bibinfo{author}{A.~Tanaka}, \bibinfo{author}{A.~D. Wade}, \bibinfo{author}{L.~M. Wagner}, \bibinfo{author}{L.~L. Wang}, \bibinfo{author}{C.~Wilhelm}, \bibinfo{author}{C.~Wu}, \bibinfo{author}{J.~Yang}, \bibinfo{author}{A.~Zamarron}, \bibinfo{author}{M.~van Zuylen}, \bibinfo{author}{D.~S. Weld},
\newblock \bibinfo{title}{{The Semantic Scholar Open Data Platform}},
\newblock \bibinfo{journal}{ArXiv} \bibinfo{volume}{abs/2301.10140} (\bibinfo{year}{2023}). \URLprefix \url{https://api.semanticscholar.org/CorpusID:256194545}.
%Type = Misc
\bibitem[{Jett et~al.(2020)Jett, Capitanu, Kudeki, Cole, Hu, Organisciak, Underwood, Dickson~Koehl, Dubnicek, and Downie}]{jett2020hathitrust}
\bibinfo{author}{J.~Jett}, \bibinfo{author}{B.~Capitanu}, \bibinfo{author}{D.~Kudeki}, \bibinfo{author}{T.~Cole}, \bibinfo{author}{Y.~Hu}, \bibinfo{author}{P.~Organisciak}, \bibinfo{author}{T.~Underwood}, \bibinfo{author}{E.~Dickson~Koehl}, \bibinfo{author}{R.~Dubnicek}, \bibinfo{author}{J.~S. Downie}, \bibinfo{title}{The {H}athi{T}rust {R}esearch {C}enter {E}xtracted {F}eatures {D}ataset (2.0)}, \bibinfo{howpublished}{HathiTrust Research Center}, \bibinfo{year}{2020}. \DOIprefix\doi{10.13012/R2TE-C227}.
%Type = Misc
\bibitem[{McCallum(2002)}]{McCallum2002}
\bibinfo{author}{A.~K. McCallum}, \bibinfo{title}{{MALLET}: A {M}achine {L}earning for {L}anguage {T}oolkit}, \bibinfo{year}{2002}. \URLprefix \url{http://mallet.cs.umass.edu}.
%Type = Article
\bibitem[{Blei et~al.(2003)Blei, Ng, and Jordan}]{blei2003latent}
\bibinfo{author}{D.~M. Blei}, \bibinfo{author}{A.~Y. Ng}, \bibinfo{author}{M.~I. Jordan},
\newblock \bibinfo{title}{Latent {D}irichlet {A}llocation},
\newblock \bibinfo{journal}{Journal of Machine Learning Research} \bibinfo{volume}{3} (\bibinfo{year}{2003}) \bibinfo{pages}{993--1022}. \URLprefix \url{https://www.jmlr.org/papers/volume3/blei03a/blei03a.pdf}.
%Type = Inproceedings
\bibitem[{Reimers and Gurevych(2019)}]{reimers-2019-sentence-bert}
\bibinfo{author}{N.~Reimers}, \bibinfo{author}{I.~Gurevych},
\newblock \bibinfo{title}{{S}entence-{BERT}: {S}entence {E}mbeddings using {S}iamese {BERT}-{N}etworks},
\newblock in: \bibinfo{booktitle}{Proceedings of the 2019 Conference on Empirical Methods in Natural Language Processing}, \bibinfo{publisher}{Association for Computational Linguistics}, \bibinfo{year}{2019}, pp. \bibinfo{pages}{3982--3992}. \URLprefix \url{https://arxiv.org/abs/1908.10084}.
%Type = Article
\bibitem[{Yin et~al.(2023)Yin, Wu, Yokota, Matsumoto, and Shibayama}]{10.1371/journal.pone.0284567}
\bibinfo{author}{D.~Yin}, \bibinfo{author}{Z.~Wu}, \bibinfo{author}{K.~Yokota}, \bibinfo{author}{K.~Matsumoto}, \bibinfo{author}{S.~Shibayama},
\newblock \bibinfo{title}{{Identify Novel Elements of Knowledge with Word Embedding}},
\newblock \bibinfo{journal}{PLOS ONE} \bibinfo{volume}{18} (\bibinfo{year}{2023}) \bibinfo{pages}{1--16}. \DOIprefix\doi{10.1371/journal.pone.0284567}.
%Type = Article
\bibitem[{Shibayama et~al.(2021)Shibayama, Yin, and Matsumoto}]{10.1371/journal.pone.0254034}
\bibinfo{author}{S.~Shibayama}, \bibinfo{author}{D.~Yin}, \bibinfo{author}{K.~Matsumoto},
\newblock \bibinfo{title}{{Measuring Novelty in Science with Word Embedding}},
\newblock \bibinfo{journal}{PLOS ONE} \bibinfo{volume}{16} (\bibinfo{year}{2021}) \bibinfo{pages}{1--16}. \DOIprefix\doi{10.1371/journal.pone.0254034}.
%Type = Article
\bibitem[{Li et~al.(2023)Li, Zhang, Zhang, Long, Xie, and Zhang}]{li2023towards}
\bibinfo{author}{Z.~Li}, \bibinfo{author}{X.~Zhang}, \bibinfo{author}{Y.~Zhang}, \bibinfo{author}{D.~Long}, \bibinfo{author}{P.~Xie}, \bibinfo{author}{M.~Zhang},
\newblock \bibinfo{title}{{Towards General Text Embeddings with Multi-Stage Contrastive Learning}},
\newblock \bibinfo{journal}{arXiv preprint arXiv:2308.03281}  (\bibinfo{year}{2023}).
%Type = Article
\bibitem[{Henderson et~al.(2017)Henderson, Al{-}Rfou, Strope, Sung, Luk{\'{a}}cs, Guo, Kumar, Miklos, and Kurzweil}]{DBLP:journals/corr/HendersonASSLGK17}
\bibinfo{author}{M.~L. Henderson}, \bibinfo{author}{R.~Al{-}Rfou}, \bibinfo{author}{B.~Strope}, \bibinfo{author}{Y.~Sung}, \bibinfo{author}{L.~Luk{\'{a}}cs}, \bibinfo{author}{R.~Guo}, \bibinfo{author}{S.~Kumar}, \bibinfo{author}{B.~Miklos}, \bibinfo{author}{R.~Kurzweil},
\newblock \bibinfo{title}{{Efficient Natural Language Response Suggestion for Smart Reply}},
\newblock \bibinfo{journal}{CoRR} \bibinfo{volume}{abs/1705.00652} (\bibinfo{year}{2017}). \URLprefix \url{http://arxiv.org/abs/1705.00652}. \href{http://arxiv.org/abs/1705.00652}{{\tt arXiv:1705.00652}}.
%Type = Misc
\bibitem[{Ouyang et~al.(2022)Ouyang, Wu, Jiang, Almeida, Wainwright, Mishkin, Zhang, Agarwal, Slama, Ray, Schulman, Hilton, Kelton, Miller, Simens, Askell, Welinder, Christiano, Leike, and Lowe}]{ouyang2022traininglanguagemodelsfollow}
\bibinfo{author}{L.~Ouyang}, \bibinfo{author}{J.~Wu}, \bibinfo{author}{X.~Jiang}, \bibinfo{author}{D.~Almeida}, \bibinfo{author}{C.~L. Wainwright}, \bibinfo{author}{P.~Mishkin}, \bibinfo{author}{C.~Zhang}, \bibinfo{author}{S.~Agarwal}, \bibinfo{author}{K.~Slama}, \bibinfo{author}{A.~Ray}, \bibinfo{author}{J.~Schulman}, \bibinfo{author}{J.~Hilton}, \bibinfo{author}{F.~Kelton}, \bibinfo{author}{L.~Miller}, \bibinfo{author}{M.~Simens}, \bibinfo{author}{A.~Askell}, \bibinfo{author}{P.~Welinder}, \bibinfo{author}{P.~Christiano}, \bibinfo{author}{J.~Leike}, \bibinfo{author}{R.~Lowe}, \bibinfo{title}{{Training Language Models to Follow Instructions with Human Feedback}}, \bibinfo{year}{2022}. \URLprefix \url{https://arxiv.org/abs/2203.02155}. \href{http://arxiv.org/abs/2203.02155}{{\tt arXiv:2203.02155}}.
%Type = Article
\bibitem[{Liu et~al.(2019)Liu, Ott, Goyal, Du, Joshi, Chen, Levy, Lewis, Zettlemoyer, and Stoyanov}]{RoBERTa2019}
\bibinfo{author}{Y.~Liu}, \bibinfo{author}{M.~Ott}, \bibinfo{author}{N.~Goyal}, \bibinfo{author}{J.~Du}, \bibinfo{author}{M.~Joshi}, \bibinfo{author}{D.~Chen}, \bibinfo{author}{O.~Levy}, \bibinfo{author}{M.~Lewis}, \bibinfo{author}{L.~Zettlemoyer}, \bibinfo{author}{V.~Stoyanov},
\newblock \bibinfo{title}{{R}o{BERT}a: {A} {R}obustly {O}ptimized {BERT} {P}retraining {A}pproach},
\newblock \bibinfo{journal}{CoRR} \bibinfo{volume}{abs/1907.11692} (\bibinfo{year}{2019}). \URLprefix \url{http://arxiv.org/abs/1907.11692}. \href{http://arxiv.org/abs/1907.11692}{{\tt arXiv:1907.11692}}.
%Type = Article
\bibitem[{Sobchuk and Šeļa(2024)}]{sobchuk2024computational}
\bibinfo{author}{O.~Sobchuk}, \bibinfo{author}{A.~Šeļa},
\newblock \bibinfo{title}{{Computational Thematics: Comparing Algorithms for Clustering the Genres of Literary Fiction}},
\newblock \bibinfo{journal}{Humanities and Social Sciences Communications} \bibinfo{volume}{11} (\bibinfo{year}{2024}) \bibinfo{pages}{438}. \DOIprefix\doi{10.1057/s41599-024-02933-6}.
%Type = Misc
\bibitem[{Griebel et~al.(2023)Griebel, Cohen, Li, Liu, Park, Perkins, and Underwood}]{griebel_cohen_li_liu_park_perkins_underwood_2023}
\bibinfo{author}{S.~Griebel}, \bibinfo{author}{R.~Cohen}, \bibinfo{author}{L.~Li}, \bibinfo{author}{J.~Liu}, \bibinfo{author}{J.~Park}, \bibinfo{author}{J.~M. Perkins}, \bibinfo{author}{W.~E. Underwood}, \bibinfo{title}{{Comparing Measures of Textual Innovation}}, \bibinfo{year}{2023}. \DOIprefix\doi{10.17605/OSF.IO/A3G6E}.

\end{thebibliography}

\appendix

\section{Topic models}

Topic granularity will vary if a corpus includes many more texts in some periods than others, and this could be problematic for a project interested in comparisons across time. So our procedure in every case was:

\begin{enumerate}
\item Restrict the corpus to an even distribution across time.

\item Generate a 250-topic model with MALLET, including an “inferencer.”
\item Use the inferencer to generate topic distributions for documents that had to be left out of the “flat” distribution in step 1.
\end{enumerate}

Using this model, we assessed novelty, transience, and precocity by measuring the K-L divergence between texts. K-L divergence is an asymmetric measure; we took the document being characterized as the reference probability distribution, and compared both past and future documents to that reference point.

\section{Embeddings}

We began by testing off-the-shelf GTE embeddings~\cite{li2023towards}. When these performed poorly, we realized that embeddings are trained mostly on twenty-first-century material, and fine-tuning would be needed to give them a better chance of representing an earlier period.

The tuning method we ultimately adopted relies on multiple negatives ranking loss, as implemented in Sentence Transformers~\cite{DBLP:journals/corr/HendersonASSLGK17, reimers-2019-sentence-bert}. That is, the training dataset includes only positive pairs of similar passages; negative pairs are created implicitly by misaligning the passages in a batch. We created positive pairs mostly by selecting adjacent passages from the same article (or work of fiction). But we adopted several tricks to prevent the model from learning a model of similarity defined purely by vocabulary overlap. First, we used GPT-3.5 to paraphrase and condense one element of some pairs~\cite{ouyang2022traininglanguagemodelsfollow}. Paraphrasing up to 18\% of pairs seemed to improve results. Second, in training embeddings for fiction, we replaced personal names in one element of each pair---preserving first and last names, and gender signals, as much as possible. Both of these changes made the learning task more difficult and improved alignment with social evidence. We used these datasets to fine-tune RoBERTa~\cite{RoBERTa2019}.

We also explored several alternate approaches that aren't represented in the final paper. For the task of predicting citations, we experimented with embeddings that were trained specifically to identify the kind of similarity between articles that produces citation. Here, positives were sentences from articles related by citation, and negatives were  pairs of sentences identified by off-the-shelf embedding methods as sharing intellectual influence, despite no documented citation existing between the two articles. Our hypothesis was that these pairs represent spurious or coincidental similarities in language not necessarily associated with the type of intellectual influence we are trying to measure. We took these pairs and fine-tuned the GTE model, through Cosine Entropy Loss, assigning high similarity to correctly identified citation pairs and low similarity to false identified pairs~\cite{li2023towards}.  

Since we were concerned that embeddings might perform less well on long passages than on individual sentences, we also tested a strategy where we generated embeddings on single sentences, then clustered them, and took the cluster centroids as synthetic “document embeddings.” This did not improve performance.

An alternate approach we have not yet checked would be to train embeddings entirely from scratch on these corpora. Some recent studies suggest that even older methods of doing that, like doc2vec, can outperform topic models on clustering tasks~\cite{sobchuk2024computational}.

We embedded passages of up to 512 tokens, with the constraint that we divide passages only at sentence breaks. Note that the chunks used for topic modeling were generally combinations of two or more embedding chunks; this difference of size was permitted in order to emphasize the strengths of both methods, without hindering either one.

\section{Text-reuse detection} 

We avoided comparing any papers written by the same author. We also aimed to avoid comparing chunks of text that directly quoted each other, as including these, we estimated, would create a circularity in the precocity calculation for such chunks, directly guaranteeing that it would correlate with citation.

To avoid this circularity, we looked for both the existence of the cited author’s last name or a string of six or more matching words that were in single or double quotations within the citing paper. If either of these are found, the chunk is not used for comparison. It is important to note that the whole paper is not excluded from comparison; only the offending chunk.

\section{Author age determination}

For the fiction corpus we could rely on previously published data to determine authors' years of birth~\cite{Underwood_2022}. 

To create analogous data for literary scholars, we estimated years of birth for a sample of 1,093 authors (and 2,646 articles) through a mixture of manual research and searches on the VIAF API. A model was trained to distinguish true VIAF matches from false ones. We estimate that we achieved overall accuracy of greater than $90\%$; this estimate is based both on the accuracy of the VIAF model and on manually checking a sample of articles.

\section{Corpus construction}

The literary studies journals included are: \emph{Publications of the Modern Language Association} (1900--2016), \emph{English Literary History} (1934--2016), \emph{The Review of English Studies} (1925--2016), \emph{Critical Inquiry} (1974--2016), \emph{Modern Language Review} (1905--2016), \emph{Modern Philology} (1903--2016), and \emph{New Literary History} (1969--2016).

The economics journals included are: \emph{The American Economic Review} (1911--2017), \emph{Econometrica} (1933--2017), \emph{Journal of Economic Literature} (1969--2017), \emph{Journal of Political Economy} (1900--2017), \emph{The Quarterly Journal of Economics} (1900--2017), \emph{The Review of Economics and Statistics} (1919--2017), \emph{The Journal of Finance} (1946--2017), and \emph{The Review of Economic Studies} (1933--2017). Both the economics and literary studies datasets were originally sourced from JSTOR~\cite{JSTOR_DfR}. Because we look back and forward 20 years in calculating precocity, we only directly characterize articles in the central period 1920--1996 (or 1997 for economics). 

For the academic articles, citations from Semantic Scholar were collected as external evidence of impact~\cite{Kinney2023TheSS}. In practice, this meant that the articles sourced from JSTOR had to be aligned with articles available in Semantic Scholar. Semantic Scholar, like all bibliographic databases, is incomplete, so all citation counts referred to in this article may underrepresent the total post-publication impact of publications. Note also that we do not limit the time period for citation, so works published earlier have, in principle, more opportunities to be cited. We address this later by controlling for date of publication. 

Our fiction corpus covers the period between 1890-2000. However, because we look back and forward 20 years in characterizing a book’s relationship to the past and future, we can only directly characterize books in the central period 1910-1979. The 20-year shoulders on either side of this period are used as comparative touchstones. So our analysis directly describes 5,880 books 1910-1979, of which we had 4,392 in full text. Since we only ran embeddings on works in full text, the smaller size of that corpus does create a slight disadvantage for the embedding method in the case of fiction.

\section{Timeline for perplexity calculation}

We calculated perplexity using RoBERTa on chunks of up to 512 tokens (the same ones we used for embedding)~\cite{RoBERTa2019}. We divided the timeline into overlapping 12-year periods with a 4-year offset, which ends up meaning that a text published in 1968-1971, for instance, would be compared to a past model trained on 1952-63 and a future model trained on 1976-87. But a text published in 1964-67 would be compared to a past model trained on 1948-59 and a future model trained on 1972-83. 

Our goal in creating 12-year models, but moving them forward 4 years at a time, was to create sufficiently large corpora for training while ensuring that texts were not greatly (dis)advantaged by their position within a time step.

\section{Domain insights}

Our primary goal in this paper is to validate a method. But it is also easy to see how this method could be used to illuminate substantive research questions about a genre or academic discipline. To give a quick sense of what it might reveal, we’ve visualized the seven journals that comprise our literary studies corpus, along with a selection of authors who have exceptionally high precocity and/or an exceptionally high number of citations.

\begin{figure}
  \centering
  \includegraphics[width=\linewidth]{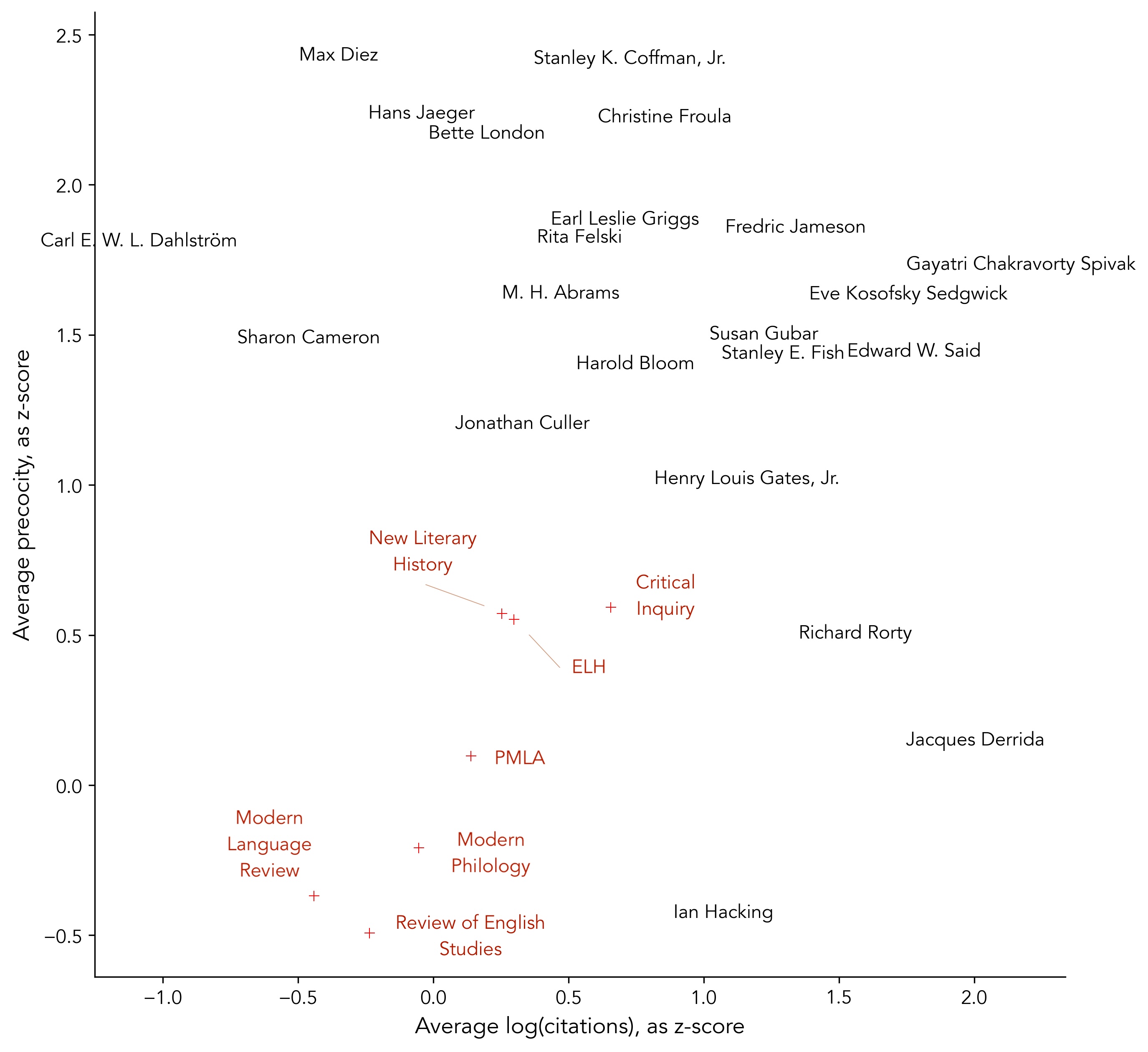}
  \caption{Literary studies journals and authors plotted by the average precocity and number of citations associated with their articles. Precocity is determined here by topic modeling. Since both axes are z-scores, the center of the whole corpus would be at 0, 0. We’re looking mostly at the upper right quadrant.}
\end{figure}

Citation counts are already public. But precocity---vertical position in Figure 1---is new information. Precocity does tend to correlate with citations, as is visible in the positive slope of the journals. But journals that attract different numbers of citations (like \emph{New Literary History} and \emph{Critical Inquiry}) may nevertheless be close to indistinguishable when it comes to precocity—which suggests they have substantively equal power to predict trends in the discipline. In other cases, journals that aren’t distinguished by citation count can be distinguished by precocity. \emph{PMLA} is the flagship journal of the Modern Language Association, and arguably the highest-prestige venue in this group. It attracts almost as many citations as \emph{New Literary History} or \emph{ELH}, but its position on the vertical axis suggests that editorial practices have sometimes been more conservative (as perhaps befits the journal of a large professional organization).

The apparent negative slope of author names is an artifact of the process we used to select exceptional authors, which deliberately highlights names on the periphery. If we plotted all authors, we would get a Gaussian cloud of points with the same slope and center as the journals (but much larger, since authors are associated with fewer articles and thus aren’t pulled to the origin as strongly by the law of averages).

The names of well-known critics, like Fredric Jameson and Gayatri Spivak, tend to be found in the upper right corner, suggesting that they were not only widely cited but prescient (or influential—causality is impossible to determine here). Moving up and to the left we find names that may be less familiar, but that our algorithm suggests were also ahead of the curve. Carl E. W. L. Dahlström is an early-twentieth-century critic whose articles have almost never been cited, although they anticipate subsequent trends.

On the right side of the graph we find a few widely-cited authors who aren’t especially distinguished by precocity. This is not necessarily a negative reflection on their work. For instance, several authors in this region (Richard Rorty, Jacques Derrida, and Ian Hacking) are well-known philosophers who were occasionally invited to publish in literary studies journals. Since they can hardly expect to convert literary scholars into philosophers en masse, these honorific late-career publications won’t stand at the beginning of a long tradition of similar work, and therefore won’t have high precocity. In short, there can be more than one kind of influence. Precocity measures a text’s relation to a specific corpus, and may not capture all the intellectual influences that flow between corpora. It is nevertheless easy to see how this metric could be used to pose questions about editorial practices and career arcs within a discipline.

\section{Preregistration and paths not taken}

Most of the methodological details above were preregistered in Fall 2023~\cite{griebel_cohen_li_liu_park_perkins_underwood_2023}. But the experimental plan did change in some important ways afterward. In particular, our embedding strategy changed several times, after off-the-shelf GTE embeddings proved not to be competitive. Also, comparison to authorial age wasn’t part of our original plan. A critical reader might (correctly) interpret these adjustments to our plan as efforts to find some method or context that would allow Transformer-based methods to outperform a topic model, as we had originally expected. If we had followed our original experimental plan exactly, the result would have been a simple endorsement of topic modeling. Evidence of our struggle to avoid or complicate that conclusion may perhaps make it even more persuasive.

There is also a question we proposed in the preregistration, and did investigate, but haven't discussed above for reasons of space. Some researchers may wonder whether it really makes sense to compare a text chunk to all the parts of all documents in the preceding and following 20 years. One could argue that mystery novels, for instance, are not really innovating relative to science fiction, but to other mystery novels. One way of taking this into account---which performed well in some previous work---was to compare chunks only to a subset of very similar chunks in the past and future (say the top 5\%)~\cite{Underwood_2022}. We also tested that strategy here, but it didn't often improve on other approaches, and so we've deferred discussion to this appendix.

\end{document}